# ShufaNet: Classification method for calligraphers who have reached the professional level

Yunfei Ge[1]   Changyu Diao[1]   Min Li[1]   Ruohan Yu[1]   Linshan Qiu [1]   Duanqing Xu[1]
[1]College of Computer Science and Technology, Zhejiang University

## Abstract

*The authenticity of calligraphy is significant but difficult task in the realm of art, where the key problem is the few-shot classification of calligraphy. We propose a novel method, ShufaNet ("Shufa" is the pinyin of Chinese calligraphy), to classify Chinese calligraphers' styles based on metric learning in the case of few-shot, whose classification accuracy exceeds the level of students majoring in calligraphy. We present a new network architecture, including the unique expression of the style of handwriting fonts called ShufaLoss and the calligraphy category information as prior knowledge. Meanwhile, we modify the spatial attention module and create ShufaAttention for handwriting fonts based on the traditional Chinese nine Palace thought. For the training of the model, we build a calligraphers' data set.*

*Our method achieved 65% accuracy rate in our data set for few-shot learning, surpassing resNet and other mainstream CNNs. Meanwhile, we conducted battle for calligraphy major students, and finally surpassed them. This is the first attempt of deep learning in the field of calligrapher classification, and we expect to provide ideas for subsequent research.*

## 1. Introduction

Chinese calligraphy is an ancient writing art of Chinese characters. For example, what shown in Figure 1 is a famous calligraphy work, "Lan Ting Xu" (Preface of the Orchid Pavilion). It was written by Wang Xizhi, a famous Chinese calligrapher. "Lan Ting Xu" is called the best running script work in the world. The task of calligrapher style classification refers to the task of judging the calligrapher who wrote these words through the analysis of single words, and the main application field of this task is the authentication of cultural relics. In the research and protection of cultural heritage, the identification of the authenticity of cultural relics is a problem that researchers must often face.

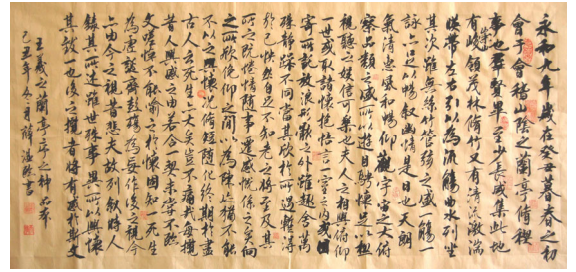

Figure 1: "Lan Ting Xu", which is called the best running script work in the world, written by Wang Xizhi.

The identification of calligraphic works in cultural heritage is particularly difficult because of their huge number and rich information. Liu deeply analyzed the value of calligraphy works in [1], so the appraisal work is particularly valuable. The classification method of traditional Chinese calligraphers is often conducted by professional calligraphers for handwriting analysis[2]. Through the handwriting, they explore its psychological process, the calligraphy works of the word or character group classification, statistics, analysis, and finally come to the conclusion of authenticity. However, the traditional method has some limitations[3]:

a) The limitation of knowledge. Traditional methods rely on the knowledge level of the calligrapher, while Chinese calligraphy has a long history. It is difficult to make a correct judgment based on human knowledge alone.

b) Unreliability of experience. Traditional methods rely too much on people, who are easily affected by subjective and objective factors. When personal mood, environmental light and other conditions change, judgment results will also be biased.

With the rapid development of deep learning, the method of deep learning has become the mainstream method of image classification and has achieved good results in many application fields. Therefore, it has become an important idea to use the method of deep learning to deal with the task of calligrapher style classification[5, 6, 7]. But there is no mature Network or architecture to deal with this problem, and few people work in the field. The main reasons lie in three aspects:

a) The calligraphy style of Chinese calligraphers is unpredictable, and its features are difficult to be explored and expressed. The existing models and methods cannot reach the level of verification of cultural relics.

b) There are many kinds of Chinese calligraphers, and their works are even more numerous. However, the existing calligraphy characters of many calligraphers are limited, which is a few-shot classification task.

c) There is no readily available data set.

In order to complete the task of calligraphy style classification, we have solved these three problems accurately, and the resulting Network model has exceeded the level of students majoring in calligraphy.

a) We crawled a large number of calligraphy characters from a number of authoritative calligraphy websites, and carried out data cleaning, and finally obtained this data set. After screening, we got 200 effective calligraphers. Besides, every character also have the label of half of calligraphy category which including Regular script, Official script, seal script, running script and cursive script. The samples can be seen in Figure 2.

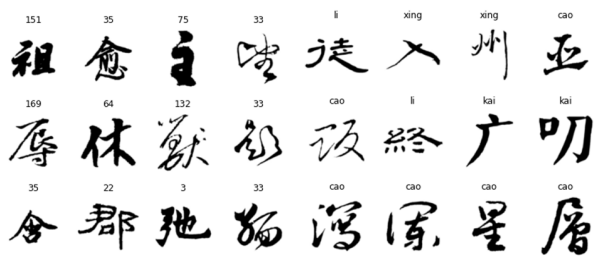

Figure 2: Sample of our dataset. Left: The label is the number of calligraphers which is ordered randomly. Right: The label is the calligraphy category. "kai" "li" "zhuan" "xing" "cao" stand for Regular script, Official script, seal script, running script and cursive script respectively.

b) Due to the wide variety of calligraphers, we built ShufaNet based on metric learning to realize the mapping of calligraphy characters into high-dimensional vectors for classification, which still have the classification ability for categories we have never seen before.

c) Style Loss based on Gram matrix[8] has achieved excellent results in the field of style transfer, but the Gram matrix only represents the overall texture features of the image, ignoring the spatial information. Based on this idea, we obtain StyleMatrix by calculating feature correlation between local spatial information, and then calculate StyleLoss according to StyleMatrix. Combined with the prior knowledge extracted from the Calligraphy Categories Network (later called CCNet), ShufaNet had a better stability. We also have modified the spatial attention module[40] and created ShufaAttention for handwriting fonts based on the traditional Chinese nine Palace thought, which will be proved be useful for improving the accuracy in the experiment.

After piecing together these modules, we obtained our ShufaNet. Through some comparative experiments, it was proved that the effect of our ShufaNet on calligrapher data set was better than that of mainstream CNNs. In order to better understand our level, we also imitated ShufaNet's training scene to conduct a questionnaire survey on the students majoring in calligraphy, and obtained the classification accuracy rate of the students majoring in calligraphy under the condition of few-shot. The experiment proved that our ShufaNet could exceed the accuracy rate and basically achieve the classification level of the students majoring in calligraphy.

In conclusion, our main contributions are as follows:

1. Create a calligrapher dataset;

2. Built ShufaNet, an architecture of calligrapher classification based on metric learning, designed ShufaAttention module and StyleMatrix, an expression way of calligraphy character style, and the classification effect exceeded that of mainstream CNNs;

3. Man-machine PK was carried out, and the final model effect reached the level of students majoring in calligraphy.

## 2. Related work

In this section, we summarize previous research related to our question. First, we do a brief survey on traditional calligraphy classification methods; Then we give a brief introduction to the development of few-shot learning. After that we introduce the development of style transfer and the loss calculation method. Finally, we will introduce the development of attention module and our improvement.

### 2.1. Traditional calligraphy classification

Traditional calligraphy classification is generally based on image features. For example, stroke extraction and corner detection are used to identify calligraphy style types in [28]. In [29, 31, 32, 33, 34, 35, 36], the style of calligraphy characters is defined based on strokes. But these methods have obvious shortcomings. Calligraphy character style is changeable, based on this method is difficult to get a good way to express the style. With the development of deep learning, related researches based on deep learning have gradually emerged. In [7, 30, 37, 38], handwriting identification, calligraphic character generation and calligraphic character style transfer are carried out based on generative adjudgment network. However, such a method is only the identification or style transfer of a certain category, and there will be a large error in the case of few-shots and multiple categories.

## 2.2. Few-shot learning

There are three general ways of few-shot learning: meta-learning [12, 22, 24], transfer learning [13, 21, 23, 25, 26] and metric learning [4, 11, 14, 15, 16, 17, 18, 19, 20, 27]. Our structure is based on the Triplet Network structure proposed by [45], which belongs to the category of metric learning. [44] proposed the concept of Siamese Network for the first time. This method has a good effect in identifying those categories that have many categories or have not been seen in the training process. Then [45] proposed Triplet Network architecture with three-headed structure based on Siamese Network. Compared with Siamese Network, negative samples were added to make the model more stable and achieve better results. In [46], it is proved that Triplet Network can achieve a good effect in face recognition and can adapt to complex environment and finally achieve 99. 63% accuracy on data set LFW. This method of learning can solve the problem of too many calligraphers, but there is still a key difference from our task. Face recognition belongs to the similarity of comparing images, while calligrapher recognition belongs to the similarity of comparing styles. The similarity of two similar calligraphy words may be very low. Therefore, direct use of Triplet Network training cannot achieve good results **(see supplementary materials)**. The key problem is to find a loss function that can measure the style of calligraphy.

## 2.3. Style transfer

In order to find the appropriate style representation, we refer to the practice of style Transfer. In [8], Gatys pioneered the separation and reorganization of image content and style by using CNNs, so as to create a brand of new artistic image. This process of transferring style information from art images to content images with the help of CNNs is called neural style transfer. Style transfer is generally based on generative adjudgment network. Gram matrix is used to measure the style of images in [8, 9, 10]. StyleLoss is used to control the style loss of generated images in this way, and good results have been achieved. Gram matrix can be understood as the eccentric covariance matrix between features, which can be used to represent the correlation between features, but the spatial information will disappear in the calculation process. As discussed in 3. 1, the traditional practice is mainly to look at the image features and spatial information of calligraphy characters, and the calculation of Gram matrix is obviously not conducive to the expression of calligraphy style. Therefore, based on this idea, we created a StyleMatrix to represent the spatial information of calligraphy characters, and calculated ShufaLoss to control the network learning, and finally achieved good results.

## 2.4. Attention module

Attention module (AM) was originally used in machine translation[39] and has now become an important concept in the field of neural networks. The initial attention model was used in natural language processing, and [40] first proposed the application of spatial attention mechanism in CV field. In SENet proposed by the author in [41], channel attention mechanism is used to obtain the importance of each channel. Then [42] proposed CBAM, which considers both spatial characteristics and channel characteristics. As discussed in 3. 3, calligraphy character features depend on spatial information expression, so spatial attention mechanism should be helpful for model optimization. Due to the pictorial nature of calligraphy characters, the effect of using the attention mechanism on them is not significant **(see supplementary material)**, which may be due to their varied styles. [47, 48] has analyzed the characteristics of the nine palaces of Chinese calligraphy, that is, calligraphers will pay attention to the space distribution of each part in the nine palaces when writing characters, that is, the importance of space for calligraphers is divided into nine regions. Based on this idea, we modify the spatial attention model and get ShufaAttention based on 9 regions (hereafter referred to as SA).

## 3. Model introduction

In this section, we will introduce our ShufaNet model in detail. First, we will introduce the key components: SA and ShufaLoss. Then the architecture of ShufaNet and training process are introduced. The architecture mainly includes CCNet of calligraphy and triplet structure of the main body.

For the sake of convenience, let's declare the use of symbols. We remember ShufaNet as $f_S, f_S(x)$ as the output feature vector of input x through ShufaNet, and $M_S^k(x)$ as the output StyleMatrix of input x in the convolution layer at the k-th layer of ShufaNet. Let $M_C^k(x)$ be the output StyleMatrix of input x at the k-th convolution layer of CCNet.

### 3.1. ShufaAttention

SA is a kind of spatial attention module [40], and the reason for doing so has been analyzed in 4. 4. Based on the nine palaces thought of traditional calligraphy, we marked the input as x, and obtained a feature vector W with length 9 through a CNN (VGGNet was used here). Then x was divided into nine blocks $x_0 \sim x_8$ with three rows and three columns, and the weighted $x_w$ was obtained by multiplying the weight of each block. The process of SA module can be seen in Figure 3.

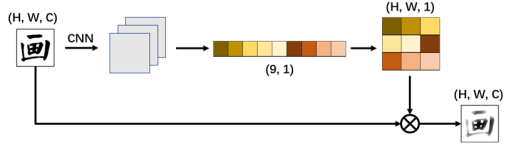

Figure 3: The process of SA module. The input image with the shape of (H, W, C) will be transformed into a feature with the length of 9 though the CNN. Finally multiply to the original input.

### 3.2. Calligraphy Category Network

In order to reduce the obvious errors(for example, regular script and running script are very different, and the classifier classified the calligrapher who belongs to regular script to another calligrapher of running script) in the training results, we introduced the prior information of calligraphy category. Calligraphy Category Network (later called CCNet) is a network that is trained end to end in the calligraphy character data set. The output of this network includes five categories, Regular script, Official script, seal script, running script and cursive script, and the intersection between this training set and ShufaNet's training set is guaranteed. Finally, the accuracy of the network is 91%, and experiments show that adding this module can increase the accuracy of the model **(see supplementary materials)**.

### 3.3. StyleMatrix and ShufaLoss

*3.3.1 StyleMatrix*

As mentioned in 3. 3, Gram matrix will pay attention to the texture information of the image and ignore the spatial position information of the image, which is the opposite for calligraphy characters. Therefore, the calculation method of our StyleMatrix is also opposite to gram matrix. We calculate the correlation between local spatial features and write the correlation matrix as StyleMatrix to represent the stylistic features of spatial information. For input x, its output at the k-th convolution layer in CNN is denoted as $output_k$. Shape of $output_k$ is $(h_k, w_k, c_k)$, $output'_k$ is obtained by flattening, with the shape of $(h_k * w_k, c_k)$, and $output'_k{}^T$ is obtained by transpose. $M^k_{CNN}(\mathbf{x})$, with the shape of $(h_k * w_k, h_k * w_k)$, is the dot product of $output'_k$ and $output'_k{}^T$. The brief computation can be seen in Figure 4.

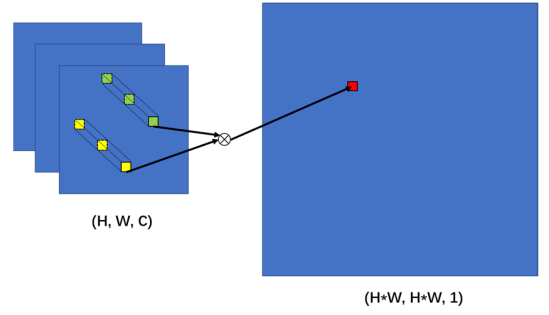

Figure 4: The computation of StyleMatrix. $M^k_{CNN}(\mathbf{x})$ is a big matrix, the element in the position of (i, j) means the correlation of the i-th and j-th spatial feature of $output_k$.

*3.3.2 ShufaLoss*

Our idea of ShufaLoss comes from style transfer[8, 9, 10], so our ShufaLoss also includes two parts, which are called StyleLoss representing style and TripletLoss representing triplet distance of the output features, and the final Loss is the sum of the two. We can calculate it using formula 1.

$$L_{shufa} = \alpha L_{style} + \beta L_{triplet} \qquad (1)$$

Since ShufaNet adopts triplet structure as a whole, The input of ShufaNet will be three images, $x_+, x$ and $x_-$, between which $x_+$ and $x$ are the characters written by the same calligrapher while $x_-$ is different. Hence, three StyleMatrix will be obtained during calculation. Here, we refer to the calculation method of tripletLoss[46] to calculate its StyleLoss . First, we get the distance of two images of the same class,

$$d_+(x_+, x) = \|x_+ - x\| \qquad (2)$$

and the distance of different classes,

$$d_-(x_-, x) = \|x_- - x\| \qquad (3)$$

After that, the tripletLoss can be calculated from the next formula, (we set margin as 200 here)

$tripletLoss(x_+, x, x_-) =$

$$\|d_+(x_+, x) - d_-(x_-, x) + margin\| \qquad (4)$$

The calculation method of StyleLoss can be seen in Figure 5. For the input of each triplet, three StyleMatrix will be obtained in each layer of CNN. The tripletLoss between them will be calculated based on Formula 4, and the value of each layer will be superimposed as the StyleLoss of the network.

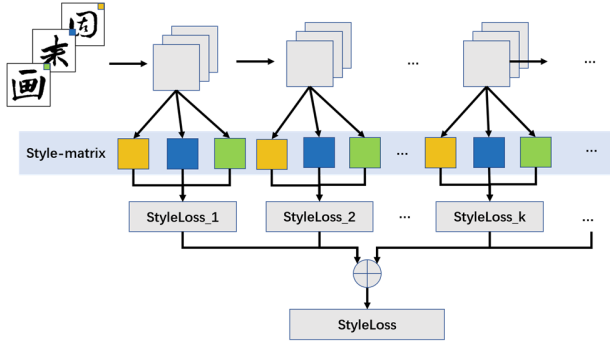

Figure 5: The calculation method of StyleLoss with the triplet input of the network.

Since the knowledge of CCNet for extracting calligraphy categories is also used in ShufaNet, the StyleLoss includes the StyleLoss through which each input goes through CCNet in addition to the theme of the Triplet schema, so the final StyleLoss should be represented as

$$L_{style} = L_{style}^{S} + L_{style}^{C} \quad (5)$$

$$= \sum_{k} tripletLoss\left(M_{S}^{k}(x_{+}), M_{S}^{k}(x), M_{S}^{k}(x_{-})\right)$$

$$+ \sum_{k} tripletLoss\left(M_{C}^{k}(x_{+}), M_{C}^{k}(x), M_{C}^{k}(x_{-})\right)$$

TripletLoss only needs to calculate the triplet structure of the subject, so its formula is

$$L_{triplet} = TripletLoss(f_s(x_+), f_s(x), f_s(x_-)) \quad (6)$$

Therefore, ShufaNet's final optimization goal is

$$\min_{f_s} \alpha L_{style} + \beta L_{triplet} \quad (7)$$

### 3.4. ShufaNet

ShufaNet is based on triplet architecture and belongs to the category of metric learning. The triplet architecture has been described in detail in related papers and will not be expanded here. Different from traditional classification, this architecture can map images to high-dimensional feature vectors, and abandon the output categories. This approach is suitable for classification with too many categories or too few samples. For categories that have never been seen before, simple training will lead to good classification ability.

The overall structure of ShufaNet is shown in the figure 6. The inputs of ShufaNet is a triplet of three images $x_+, x$ and $x_-$. $x_+$ and $x$ are similar (abbreviated by the same calligrapher), $x$ and $x_-$ are different (written by different calligraphers). First of all, the three images are obtained by SA respectively and we get the weighted images $x_+^w, x^w$ and $x_-^w$. Then, the weighted images are input into the triplet structure for training, and their StyleLoss will be calculated according to the formula introduced in 3. 3. At the same time, three images will be input into CCNet to obtain their calligraphy category features, and the StyleLoss of these images in CCNet is obtained. The Loss is superimposed into the StyleLoss of the main triplet structure. This will increase the distance between the feature vectors of different calligraphy categories and reduce the classification error. Finally, the triplet outputs three feature vectors, calculates the TripletLoss between the output feature vectors. Finally, we calculate the weighted sum of StyleLoss to get the final Loss. After the network training, a CNN that can map calligraphy characters into high-dimensional vectors is obtained. After that a classifier is trained to classify the obtained feature vectors.

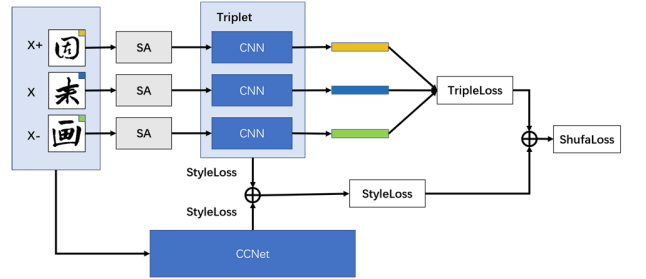

Figure 6: The architecture of ShufaNet.

## 4. Experiment

In this section, we conduct the comparative training of ShufaNet and traditional ImageNet classic network on our calligraphy character data set, including the introduction of data set, baseline, ShufaNet, questionnaire, other details of implementation and analysis. In addition, we conducted different shot experiments to further explore the superiority of ShufaNet and other CNNs. Finally, the effect of ShufaNet was analyzed based on Class Activation Mapping (CAM)[43] visualization technique.

### 4.1. Data set

In this experiment, the data set is the calligraphy character data set collected by us. We have collected calligraphy characters of 200 ancient calligraphers. At the same time we recorded the calligraphy categories. For example, the running script of Zhizhang He, a famous ancient calligrapher, and the cursive script of him are labeled as different classes. In order to simulate the scene of few-shot learning, the sample size of the dataset is not balanced, but is distributed in a step-reduction way from 2000 to 20. For convenience, this data set will be referred to as $S$ later. What's more, unless otherwise noted, the validation set is 20% of the data set.

For ShufaNet training and ImageNet network training, the data set is divided differently. For ShufaNet, as it belongs to a few shot training, the data set is randomly divided into $S_t$ and $S_{query}$. For $S_t$, it is randomly divided into two parts, $S_1$ and $S_2$. $S_1$ is used to train parameters of triplet, while $S_2$ is used to train CCNet with labels of calligraphy category. After the training is complete, the $S_{query}$ will conduct the few-shot training and verification. For the training of VGGNet16, resNet19 and resNet34, we process directly on the full data set $S$.

### 4.2. Baseline

Since ShufaNet is a new architecture and our data set is brand new, there are no existing results for comparison. In order to measure the effect of ShufaNet, we compared with some mainstream CNNs, including VGGNet16, resNet19 and resNet34. Firstly, data set $S$ is randomly divided into training set and valid set, and end-to-end training is carried out on the training set. To be fair, each network model was trained by 20 epochs. It should be noted that baseline training is not a few-shot task, but rather the result of training on a large data set with 200 classes. The loss function of all of them is CrossEntropyLoss and other details will be introduced in 4. 5.

### 4.3. ShufaNet

The training of ShufaNet includes CCNet pre-training and triplet architecture training. The first is the pre-training of CCNet. The pre-training of CCNet is based on VGGNet16, and the data set is $S_2$ randomly divided from $S_t$. Considering only the labels of its calligraphy category, 5 categories (Regular script, Official script, seal script, running script and cursive script) are finally obtained, among which the data amount of each category is about 2000. CrossEntropyLoss is used as the loss function. After the training of 20 epochs, the CNN is extracted for the training of triplet architecture. The triplet architecture is trained on $S_1$. Since the same word is certainly more similar, we introduce two types of triplets for the Net to learn about styles rather than glyphs,

$triplet_1 = \{ (x_+, x, x_-) \mid x_+, x, x_- \in S_1 \ \&\& \ x_+, x \ is \ not$

$the \ same \ word \ \&\& \ x, x_- \textbf{ is } the \ same \ word\}$

$triplet_2 = \{ (x_+, x, x_-) \mid x_+, x, x_- \in S_1 \ \&\& \ x_+, x \ is \ not$

$the \ same \ word \ \&\& \ x, x_- \textbf{ is not } the \ same \ word\}$

The training set of $S_1$ is the set of the two,

$trainingSet = \{ y_1, y_2 \mid y_1 \in triplet_1, y_2$
$\in triplet_2 \ \&\& \ |y_1| \ equlas \ |y_2| \}$

Finally, the sum number of $triplet_1$ and $triplet_2$ is 240808, with the proportion of 1:1, which were selected from $S_1$. Then, a total of 50, 000 batches were trained using 32 triples as a batch. The experiment adopts two ways, one adding SA module, the other not adding SA module, to explore the effect of SA. For each input, input to the Triplet structure is entered simultaneously into CCNet to compute its StyleLoss, and Loss is calculated based on the method in 3. 3. After the network converges, $S_{query}$ will be input into the network to obtain its feature vector, and then a classifier is trained to classify its feature vector. Finally, the classification accuracy rate in $S_{query}$ is obtained as ShufaNet's accuracy rate. $S_{query}$ is a subset randomly divided from $S$, which includes ten classes. 20 samples are divided from each class for training, and the rest are used as valid set to calculate the accuracy. CrossEntropyLoss is used as the loss function, and it is also trained for 20 epochs. What's more, we have conducted 10-shot and 5-shot learning to further explore the superiority of ShufaNet and other CNNs.

### 4.4. Questionnaire

As introduced in the first section, our ultimate goal is to achieve computer-based calligraphy character identification technology. The ultimate goal of this technology is to exceed the accuracy rate of calligraphy experts. Therefore, we also arrange a questionnaire survey to count the accuracy rate of students majoring in calligraphy. The questionnaire simulated the training process of 10 ways few-shot learning. 10 calligraphers were randomly selected from our data set. After the training set was selected, the remaining data were disrupted and matched by students as what the ShufaNet has done in training. Finally, a total of 25 questionnaires were collected, among which 19 were valid, which is about the number of new calligraphy majors in a university each year.

### 4.5. Other implementation details

The experimental computer system was Ubuntu2010, and the hardware configuration is Quadro P5000 with the memory of 16GB and 128GB RAM. Python3. 8. 3, pytorch 1. 7. 1 and fastaiv2 are used as the training architecture for deep learning. The architecture of CCNet is VGGNet16, CCNet and baseline are based on Fast. ai 2. 0, and the learning rate is calculated according to lr_find function in Fast. ai 2. 0. The backbone of ShufaNet is VGGNet16 and batch size is 32. The learning rate is initialized to 0. 0001, and then adjusted each time after 2000 batches, finishing the training of 50000 batches. Few-shot learning of ShufaNet on $S_{query}$ is also based on the Fast. ai 2. 0 architecture with the same config as baselines.

### 4.6. Analysis

#### 4.6.1 Results

First of all, the CCNet get the accuracy of 91%. The confusion matrix of CCNet can be seen in Figure 7.

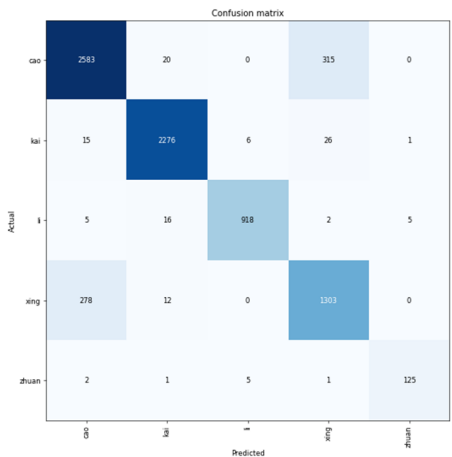

Figure 7: The confusion matrix of CCNet.

It can be seen from the figure that almost all errors are in the classification of cursive script and running script, because cursive script and running script are relatively similar when looking only at the glyph. What's more, there is even a kind of calligraphy classification named running cursive script. In a word, CCNet has been able to extract the general feature of calligraphy categories.

After the CCNet, the comparison of the accuracy, train loss and valid loss between ShufaNet and the baselines are shown from Figure 8 to Figure 10.

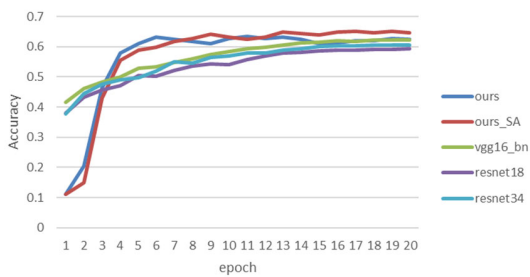

Figure 8: The accuracy of ShufaNet and baselines. Among them, ours means the ShufaNet that doesn't use SA module and ours_SA means that use SA module.

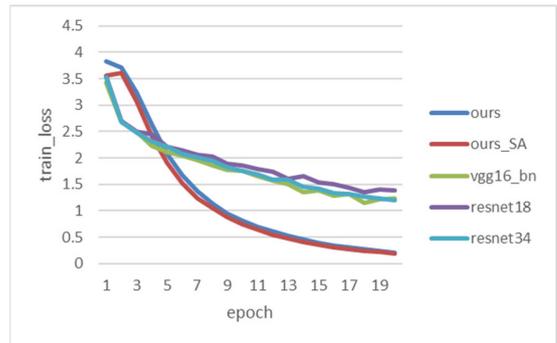

Figure 9: The train loss of ShufaNet and baselines.

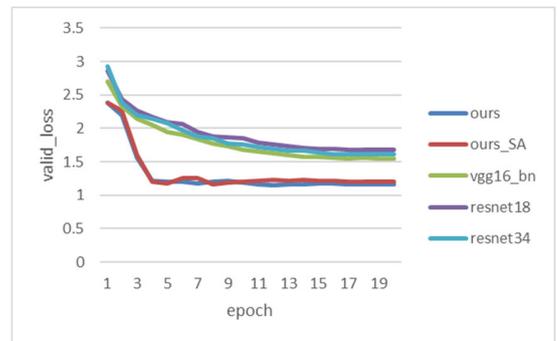

Figure 10: The valid loss of ShufaNet and baselines.

It can be seen from the figures that in the first 4 epochs, both the accuracy rate and train loss of our methods are lagging behind. This is because our method is based on metric learning. In the early stage of learning, we only map images into feature vectors, but the ability of classification is weak. However, when adapting to the distribution of the new data set, the accuracy rate began to rise rapidly and exceeded the baseline, and loss also decreased to a lower level. We can also see from the results that the effect without SA is better than VGGNet16, followed by resNet, and the accuracy rate can be improved by 2% after SA is added. From the comparison of train loss curve and valid loss curve, we can also see that ShufaNet has better stability and better robustness in expressing complex and changeable calligraphy features. From the figures we also can see that VGGNet performs better than ResNet in the network of style classes. The reason may be that the feature distribution of style classes has large variance and poor stability, so the representation of more complex networks will be easier to overfit. On the whole, the accuracy rate is only about 60%, which is far behind other classification tasks of ImageNet. The reason is that previous tasks tend to compare the content similarity of two images, while style is a higher dimensional feature, so it is difficult to get better results.

Finally, in Table 1 we can see the all result including baselines, ShufaNet of 5-shot, 10-shot and 20-shot, and the result of questionnaire.

| Method | Accuracy |
|---|---|
| **ShufaNet-SA-20-shot（ours）** | **64. 58 ± 0. 37** |
| ShufaNet-SA-10-shot | 54. 88 ± 0. 16 |
| ShufaNet-SA-5-shot | 41. 58 ± 0. 42 |
| VGGNet16 | 62. 17 ± 0. 09 |
| resNet19 | 59. 29 ± 0. 26 |
| resNet34 | 60. 62 ± 0. 03 |
| questionnaire | 61. 11 |

Table 1: The results of ShufaNet, baselines and questionnaire. The best method and its accuracy has been bolded.

From the table, we can see that our method is better than the baseline methods. With the increase of shot number, the accuracy will increase, and the accuracy will increase about 10% for doubling the sample number. The accuracy rate of students majoring in calligraphy is about 60%, and our method exceeds the average level of students majoring in calligraphy.

### 4.6.2 Others

In order to better understand the improvement of ShufaNet, we visualized the model attention based on Class Activation Mapping (CAM) visualization technique. We visualized the network of VGGNet16 and the ShufaNet-SA-20-shot model, and finally compared the effect picture as shown in Figure 11.

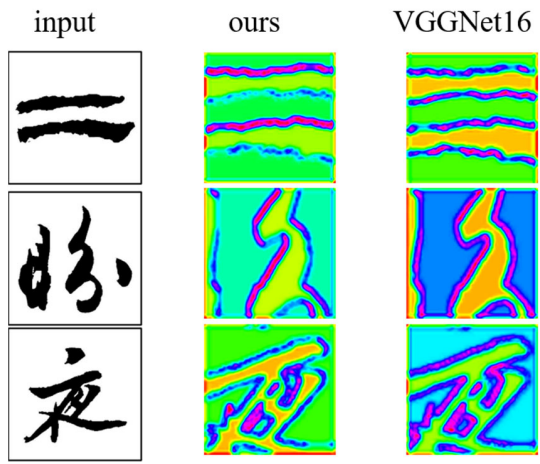

Figure 11: The Class Activation Mapping (CAM) visualization of several inputs.

It can be seen from the figure, for the outline of the same word, VGGNet16 directly simply focus on its contour shape on both sides of the strokes, and ShufaNet tend to focus on the one side. According to the writing style of calligraphy, it can be seen that this side is generally the force side of brush contact. This side can better reflect the characteristics of a calligrapher's power and his writing habits, for which it get better result. The reason may be that StyleMatrix calculates the correlation between spatial information, so that more important spatial information can be found, so as to reduce the interference of useless information and ultimately improve the accuracy.

## 5. Conclusion

It is a difficult task to identify calligraphers according to calligraphic characters, and a good model needs the ability to learn from few-shot learning. Based on the triplet structure, we proposed ShufaNet that achieves the learning ability of few-shot. Meanwhile, inspired by style transfer, we defined StyleMatrix, the expression of calligraphic character style. Based on this method and combined with triplet architecture, we defined the calculation method of ShufaLoss. What's more, we combined the prior knowledge of calligraphy category in ShufaNet and created the ShufaAttention module based on the nine Palace thought of Traditional Chinese calligraphy. Finally, we reached the level of students majoring in calligraphy. We have proved that this model outperforms some classical ImageNet models in few-shot training on calligraphy classification data set through comparative experiments. We believe that our method has great reference significance for metric learning and feature fusion in the field of handwriting fonts. Meanwhile, our method is still instructive for the classification tasks of coarse-grained and fine-grained [49], and we hope we can provide ideas for other's work. In the future, we will continue to optimize the method, try other new and innovative ways, and strive to reach the classification level of professional calligraphers.

# References

[1] Kaiyun L I U. Seek the Tension between Painting and Calligraphy Value and Market Value——Inspiration on Seeing Zhen Banqiao's Painting'Ad. [J]. Southeast Culture, 2005, 6.

[2] Latash M L , F Danion, Scholz J F , et al. Approaches to analysis of handwriting as a task of coordinating a redundant motor system[J]. Human Movement Science, 2003, 22(2):153-171.

[3] Sheng S Y . The Development of Chinese Calligraphy in Relation to Buddhism and Politics during the Early Tang Era[J]. Dissertations & Theses - Gradworks, 2011, 51(27):7980-8.

[4] Tay Y, Anh Tuan L, Hui S C. Latent relational metric learning via memory-based attention for collaborative ranking[C]//Proceedings of the 2018 World Wide Web Conference. 2018: 729-739.

[5] Lecoutre A, Negrevergne B, Yger F. Recognizing art style automatically in painting with deep learning[C]//Asian conference on machine learning. PMLR, 2017: 327-342.

[6] Zhuang Y, Lu W, Wu J. Latent Style Model: Discovering writing styles for calligraphy works[J]. Journal of Visual Communication and Image Representation, 2009, 20(2): 84-96.

[7] Yang S, Wang Z, Wang Z, et al. Controllable artistic text style transfer via shape-matching gan[C]//Proceedings of the IEEE/CVF International Conference on Computer Vision. 2019: 4442-4451.

[8] Gatys L A, Ecker A S, Bethge M. Image style transfer using convolutional neural Networks[C]//Proceedings of the IEEE conference on computer vision and pattern recognition. 2016: 2414-2423.

[9] Jin H, Wang T, Zhang M, et al. Neural Style Transfer for Picture with Gradient Gram Matrix Description[C]//2020 39th Chinese Control Conference (CCC). IEEE, 2020: 7026-7030.

[10] Zheng Z , Liu J . P^2-GAN: Efficient Style Transfer Using Single Style Image[J]. 2020.

[11] Allen K, Shelhamer E, Shin H, et al. Infinite mixture prototypes for few-shot learning[C]//International Conference on Machine Learning. PMLR, 2019: 232-241.

[12] Kaidi Cao, Maria Brbic, and Jure Leskovec. Concept learners for few-shot learning. InProc. International Conferenceon Learning Representations (ICLR), 2021.

[13] Guneet Singh Dhillon, Pratik Chaudhari, Avinash Ravichan-dran, and Stefano Soatto. A baseline for few-shot image clas-sification. InInternational Conference on Learning Repre-sentations, 2019.

[14] Carl Doersch, Ankush Gupta, and Andrew Zisserman. Crosstransformers: spatially-aware few-shot transfer. InAd-vances in Neural Information Processing Systems (NeurIPS), 2020.

[15] Gidaris S, Bursuc A, Komodakis N, et al. Boosting few-shot visual learning with self-supervision[C]//Proceedings of the IEEE/CVF International Conference on Computer Vision. 2019: 8059-8068.

[16] Spyros Gidaris and Nikos Komodakis. Generating classifi-cation weights with gnn denoising autoencoders for few-shotlearning. InProc. IEEE Conference on Computer Vision andPattern Recognition (CVPR), 2019

[17] Ruibing Hou, Hong Chang, MA Bingpeng, Shiguang Shan, and Xilin Chen. Cross attention Network for few-shot clas-sification. InAdvances in Neural Information ProcessingSystems (NeurIPS), 2019

[18] Hongyang Li, David Eigen, Samuel Dodge, Matthew Zeiler, and Xiaogang Wang. Finding task-relevant features for few-shot learning by category traversal. InProc. IEEE Confer-ence on Computer Vision and Pattern Recognition (CVPR), 2019

[19] Wenbin Li, Lei Wang, Jinglin Xu, Jing Huo, Yang Gao, andJiebo Luo. Revisiting local descriptor based image-to-classmeasure for few-shot learning. InProc. IEEE Conference onComputer Vision and Pattern Recognition (CVPR), 2019.

[20] Yann Lifchitz, Yannis Avrithis, and Sylvaine Picard. Localpropagation for few-shot learning. InInternational Confer-ence on Pattern Recognition (ICPR), 2021.

[21] Bin Liu, Yue Cao, Yutong Lin, Qi Li, Zheng Zhang, Ming-sheng Long, and Han Hu. Negative margin matters: Under-standing margin in few-shot classification. InProc. Euro-pean Conference on Computer Vision (ECCV), 2020.

[22] Puneet Mangla, Nupur Kumari, Abhishek Sinha, MayankSingh, Balaji Krishnamurthy, and Vineeth N Balasubrama-nian. Charting the right manifold: Manifold mixup for few-shot learning. InIEEE Winter Conference on Applications ofComputer Vision (WACV), 2020.

[23] Juhong Min, Dahyun Kang, and Minsu Cho. Hypercorrela-tion squeeze for few-shot segmentation. InProc. IEEE In-ternational Conference on Computer Vision (ICCV), 2021

[24] Mengye Ren, Eleni Triantafillou, Sachin Ravi, Jake Snell, Kevin Swersky, Joshua B Tenenbaum, Hugo Larochelle, andRichard S Zemel. Meta-learning for semi-supervised few-shot classification. InProc. International Conference onLearning Representations (ICLR), 2018.

[25] Yonglong Tian, Yue Wang, Dilip Krishnan, Joshua B Tenen-baum, and Phillip Isola. Rethinking few-shot image classifi-cation: a good embedding is all you need? InProc. EuropeanConference on Computer Vision (ECCV), 2020.

[26] Imtiaz Ziko, Jose Dolz, Eric Granger, and Ismail Ben Ayed. Laplacian regularized few-shot learning. InProc. Interna-tional Conference on Machine Learning (ICML), 2020

[27] Chi Zhang, Yujun Cai, Guosheng Lin, and Chunhua Shen. Deepemd: Few-shot image classification with differentiableearth mover's distance and structured classifiers. InProc. IEEE Conference on Computer Vision and Pattern Recogni-tion (CVPR), 2020

[28] Wu Y, Zhuang Y, Pan Y, et al. Web based chinese calligraphy learning with 3-d visualization method[C]//2006 IEEE International Conference on Multimedia and Expo. IEEE, 2006: 2073-2076.

[29] Yang L, Li X. Animating the brush-writing process of Chinese calligraphy characters[C]//2009 Eighth IEEE/ACIS International Conference on Computer and Information Science. IEEE, 2009: 683-688.

[30] Lyu P, Bai X, Yao C, et al. Auto-encoder guided GAN for Chinese calligraphy synthesis[C]//2017 14th IAPR International Conference on Document Analysis and Recognition (ICDAR). IEEE, 2017, 1: 1095-1100.

[31] Zhang X, Liu G. Chinese calligraphy character image synthesis based on retrieval[C]//Pacific-Rim Conference on Multimedia. Springer, Berlin, Heidelberg, 2009: 167-178.

[32] Zhang J, Lin H, Yu J. A novel method for vectorizing historical documents of Chinese calligraphy[C]//2007 10th IEEE International Conference on Computer-Aided Design and Computer Graphics. IEEE, 2007: 219-224.
[33] Zhang X, Zhao Q, Xue H, et al. Interactive creation of Chinese calligraphy with the application in calligraphy education[M]//Transactions on Edutainment V. Springer, Berlin, Heidelberg, 2011: 112-121.
[34] Wang X, Liang X, Sun L, et al. Triangular mesh based stroke segmentation for chinese calligraphy[C]//2013 12th International Conference on Document Analysis and Recognition. IEEE, 2013: 1155-1159.
[35] Zhu X W, Yang C Q. Graph Based Stroke Extraction for Chinese Calligraphy[J]. Software Guide, 2019.
[36] Lin G, Guo Z, Chao F, et al. Automatic stroke generation for style-oriented robotic Chinese calligraphy[J]. Future Generation Computer Systems, 2021, 119: 20-30.
[37] Wu R, Zhou C, Chao F, et al. GANCCRobot: Generative adversarial Nets based chinese calligraphy robot[J]. Information Sciences, 2020, 516: 474-490.
[38] Lian Z, Zhao B, Xiao J. Automatic generation of large-scale handwriting fonts via style learning[M]//SIGGRAPH Asia 2016 Technical Briefs. 2016: 1-4.
[39] Chorowski J, Bahdanau D, Cho K, et al. End-to-end Continuous Speech Recognition using Attention-based Recurrent NN: First Results[J]. Eprint Arxiv, 2014.
[40] Wu L, Wang Y. Where to Focus: Deep Attention-based Spatially Recurrent Bilinear Networks for Fine-Grained Visual Recognition[J]. 2017.
[41] Hu J, Shen L, Sun G. Squeeze-and-excitation Networks[C]//Proceedings of the IEEE conference on computer vision and pattern recognition. 2018: 7132-7141.
[42] Woo S, Park J, Lee J Y, et al. Cbam: Convolutional block attention module[C]//Proceedings of the European conference on computer vision (ECCV). 2018: 3-19.
[43] Zhou B, Khosla A, Lapedriza A, et al. Learning deep features for discriminative localization[C]//Proceedings of the IEEE conference on computer vision and pattern recognition. 2016: 2921-2929.
[44] Chopra S, Hadsell R, LeCun Y. Learning a similarity metric discriminatively, with application to face verification[C]//2005 IEEE Computer Society Conference on Computer Vision and Pattern Recognition (CVPR'05). IEEE, 2005, 1: 539-546.
[45] Hoffer E, Ailon N. Deep metric learning using triplet Network[C]//International workshop on similarity-based pattern recognition. Springer, Cham, 2015: 84-92.
[46] Schroff F, Kalenichenko D, Philbin J. FaceNet: A unified embedding for face recognition and clustering[C]//Proceedings of the IEEE conference on computer vision and pattern recognition. 2015: 815-823.
[47] Han C C, Chou C H, Wu C S. An interactive grading and learning system for chinese calligraphy[J]. Machine Vision and Applications, 2008, 19(1): 43-55.
[48] Zhang Z, Wu J, Yu K. Chinese calligraphy specific style rendering system[C]//Proceedings of the 10th annual joint conference on Digital libraries. 2010: 99-108.
[49] Bukchin G, Schwartz E, Saenko K, et al. Fine-grained Angular Contrastive Learning with Coarse Labels[C]//Proceedings of the IEEE/CVF Conference on Computer Vision and Pattern Recognition. 2021: 8730-8740.